\documentclass{esannV2}
\usepackage{graphicx}
\usepackage[latin1]{inputenc}
\usepackage{amssymb,amsmath,array}
\usepackage{multirow}
\usepackage{rotating}
\usepackage{algorithmic}
\usepackage[ruled,linesnumbered]{algorithm2e}
\usepackage{rotating}
\usepackage{booktabs} 
\usepackage{lineno}
\usepackage{caption}
\usepackage{subcaption}

\begin{document}
\title{On Feature Selection Using Anisotropic General Regression Neural Network}

\author{Federico Amato$^1$, Fabian Guignard$^1$, Philippe Jacquet$^2$ and Mikhail Kanevski$^1$
%
\thanks{F. Guignard and M. Kanevski acknowledge the support of the National Research Program 75 "Big Data" (PNR75, project No. 167285 "HyEnergy") of the Swiss National Science Foundation (SNSF).}
%
\vspace{.3cm}\\
%
1- University of Lausanne - Faculty of Geosciences and Environment, IDYST \\
UNIL, 1015 Lausanne - Switzerland
\\
2- Scientific Computing and Research Support Unit, Computer Center, \\
UNIL, 1015 Lausanne - Switzerland
}

\maketitle

\begin{abstract}
The presence of irrelevant features in the input dataset tends to reduce the interpretability and predictive quality of machine learning models. Therefore, the development of feature selection methods to recognize irrelevant features is a crucial topic in machine learning.
Here we show how the General Regression Neural Network used with an anisotropic Gaussian Kernel can be used to perform feature selection. 
A number of numerical experiments are conducted using simulated data to study the robustness of the proposed methodology and its sensitivity to sample size.  Finally, a comparison with four other feature selection methods is performed on several real world datasets.
\end{abstract}

\section{Introduction}

Machine Learning (ML) approaches have become popular tools to analyze, model and extract knowledge from data and to understand complex non-linear phenomena. However, because of the previously unseen growth of data collection and storage, recognizing the important predictors associated with response variables among a large set of features has become an extremely challenging task \cite{Guyon}. The presence of irrelevant features in a dataset amplifies the effects of the well-known curse of dimensionality \cite{Verleysen}.

Feature Selection (FS) has therefore become a crucial issue. Features can be broadly classified into completely irrelevant, weakly relevant and redundant, weakly relevant and non-redundant, or strongly relevant \cite{yu2004efficient}. A robust FS algorithm will select the features belonging to the last two categories, generally named as relevant.
In this paper we will show how Anisotropic General Regression Neural Network (AGRNN), proposed by Specht \cite{SpechtAGRNN} as an adaptation of the Nadaraya-Watson estimator \cite{nadaraya1964estimating} for prediction, can be used to perform FS when used with an anisotropic Gaussian Kernel. Two experimental case studies based on simulated and real data, respectively, are discussed in detail. 

The remainder of the paper is organized as follows. Section \ref{NW}  briefly presents the AGRNN. Section \ref{FS} describes the novel FS algorithm. Section \ref{experimental} presents the experimental results. Section \ref{software} gives information about the software availability. Finally, Section \ref{conclusions} presents some conclusions and gives a plan for the future work.

\section{The General Regression Neural Network}\label{NW}

The General Regression Neural Network (GRNN) is a regression model. Theoretically, the regression of a scalar $y$ on a vector of independent variables $\textbf{x}$, minimizing mean square error, is defined as a conditional mean value:

\begin{equation*}
    E_{Y|X}= \frac{\int_{-\infty}^{\infty}y\:f_{XY}(\textbf{x},y)\:dy}{\int_{-\infty}^{\infty}f_{XY}(\textbf{x},y)\: dy} 
\end{equation*}

However, in real case studies the joint probability density function $f_{XY}(\textbf{x},y)$ is not known. Though, it can be approximated using a multivariate Parzen-Rosenblatt estimator. For a set of $n$ sample observations this will result in the fundamental equation of the GRNN:

\begin{equation*}
    \hat{y}(\textbf{x})= \hat{f}(\textbf{x})= \frac{\sum_{i=1}^{n}y_i\: K \: \big( \frac{\textbf{x} - \textbf{x}_i}{h}\big) }{\sum_{i=1}^{n} K \: \big( \frac{\textbf{x} - \textbf{x}_i}{h}\big)}
\end{equation*}
where $K(\cdot)$ is a kernel function  and \textit{h} is its smoothing parameter. As an interpretation, $\hat{y}(\textbf{x})$ can be considered as a weighted average of all the observed values $y_i$, depending on the distance from the measurement of $\textbf{x}$.

\section{Feature Selection Using Anisotropic General Regression Neural Network}\label{FS}

AGRNN is an evolution of the GRNN in which an anisotropic Gaussian kernel function $K(\cdot)$ with a different bandwidth for each feature is used \cite{SpechtAGRNN}. Hence, the kernel assumes the form:

\begin{equation*}
K \: \bigg( \frac{\textbf{x} - \textbf{x}_i}{h}\bigg) = \exp \Bigg( \sum_{j=1}^d - \bigg( \frac{(x_j - x_{ij})^2}{2\sigma_j ^2}\bigg) \Bigg)
\end{equation*}
where an Euclidean distance between points is applied and $\sigma_j$ is the bandwidth of the Gaussian kernel for the $j^{th}$ dimension.

A proper calibration of the bandwidths will scale the input features depending on their explanatory power. When  the bandwidth of the $l^{th}$ variable grows, its contribution to the regression function will tend to zero. Conversely, a small smoothing parameter will give rise to a high discriminative power of the associated feature. This behaviour appears more clearly when considering the AGRNN as a product of exponential kernels:

\begin{equation*}
\lim_{\sigma_l\to\infty} \prod_{j=1}^p \exp \bigg[ -\bigg( \frac{(x_j - x_{ij})^2}{2\sigma_j ^2}\bigg) \bigg] = \prod_{\substack{j=1 \\ j\ne l}}^p \exp \bigg[ -\bigg( \frac{(x_j - x_{ij})^2}{2\sigma_j ^2}\bigg) \bigg]
\end{equation*}

Considering this property, we can define a criterion to select relevant features with the use of AGRNN as follows. Let $\mathcal{X}=\{X_1,X_2,\dots,X_j, \dots, X_d\}$ be the set of $d$ variables constituting an input space and $Y$ be its corresponding response. Given a vector $\Sigma=(\sigma_1,\sigma_2, \dots, \sigma_j, \dots, \sigma_d )$ of the bandwidths minimizing the loss function of an AGRNN, we say that the $j^{th}$ variable is relevant if $\sigma_j $ is smaller than or equal to a specific threshold, here fixed at 1 in the case in which the features are normalized to the interval $[0,1]$. 
Indeed, when $\sigma_j > 1$ the weights defined by the Gaussian kernel will tend to assume the same small value for all the training points, and the contribution of the $j^{th}$ feature to the regression will become negligible. 

The pseudocode for the AGRNN based Selector (AS) proposed in the previous lines is described in Algorithm 1.

\begin{algorithm}[H]
 \SetAlgoLined
 \SetKwInOut{Input}{Input}\SetKwInOut{Output}{Output}
 
 \Input{A dataset $\mathcal{X}$ with features $\{X_1,X_2,\dots,X_j, \dots, X_d\}$ and its corresponding response $Y$.}
 \Output{A vector $\Sigma=(\sigma_1,\sigma_2, \dots, \sigma_j, \dots, \sigma_d )$ of the bandwidths; a dataset $\widetilde{\mathcal{X}}$ containing the relevant features.}

 Rescale each feature to [0,1].\;
 
 Train an AGRNN to compute the optimal $\Sigma$.
 
 \For{i = 1 \textbf{to} d \do}{
    \If{$\sigma_j \leq 1$}{
    Store $X_i$ in $\widetilde{\mathcal{X}}$\;
    }
 }
 \caption{AS}
\end{algorithm}

Clearly, the identification of relevant feature through this AS is sensitive to the training of the model, i.e. to the identification of proper bandwidth values. In this research the Limited-memory BFGS method \cite{nocedal1980updating} has been used to solve the $\sigma_j$ optimization problem. A strong advantage of the AS compared to other FS methods is that it studies all the features at once, considering their non-linear interactions. Moreover, the different values of sigma can give an indication concerning the relative importance of each feature. Although in this paper it is used only to discriminate irrelevant features, the AGRNN can be used to solve the regression problem. In this case, the value of the bandwidths acts as a weight given to each feature, and irrelevant variables are automatically filtered out from the regression as shown in the previous equations. At the same time, weakly relevant features, either redundant or not, will be given higher bandwidth values than the relevant features. Because of this peculiar behaviour, AGRNN can also be considered as an embedded FS method. 

\section{Experimental Study}\label{experimental}

In this section we explore the behaviour of AS with two experimental case studies with simulated and real datasets, respectively. All the computations have been executed on an Intel(R) core i7-8700K 3.70GHz with 32 GB of RAM.

\subsection{Synthetic data}

The simulated \textit{Butterfly} dataset, introduced in \cite{Golay2017}, is constituted by one target variable $Y$ generated from two uniformly distributed inputs $X_1, X_2 \in ]-5,5[$ by using a neural network with one hidden layer of 10 neurons and fixed weights. Three more features are then generated as a combination of the two relevant features: $J_3 = \log_{10}(X_1+5)$, $J_4=X_1^2-X_2^2$ and $J_5=X_1^4-X_2^4$. Finally, three irrelevant features are added: $I_6 \in ]-5,5[$ randomly sampled from a uniform distribution, $I_7=\log_{10}(I_6+5)$ and $I_8=I_6+I_7$.

\begin{figure}[h]
\centering\includegraphics[width=1\linewidth]{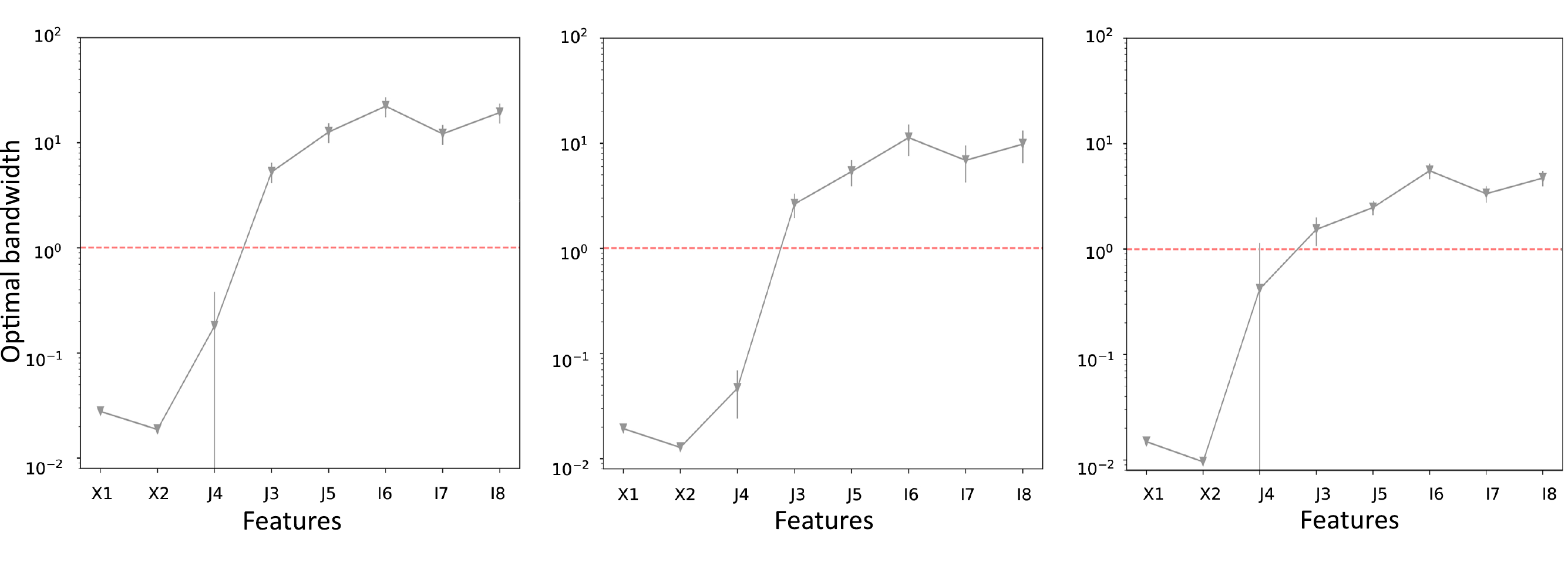}
\caption{Mean and  95\% confidence interval for the optimal bandwidth values for each feature after the optimization with the Limited-memory BFGS algorithm over 100 generated datasets for different sample sizes (from left $n = 2000$, $n = 5000$, $n = 10000$). The red dashed line represents the threshold used for the FS. Notice that the ordinate is in log-scale.}
\label{fig:sigmas}
\end{figure}

To study this dataset with the AS, 100 simulations were generated with different sample size, specifically $n=2000$, $n=5000$ and $n=10000$ and used to test the sensitivity of the AS to the number of training points. For each sample size AS has been performed to find the optimal bandwidths $\sigma_{opt}$ for each feature on each of the 100 generated datasets. Figure \ref{fig:sigmas} shows the mean and the confidence interval for the obtained bandwidths over the 100 repetitions for the three investigated sample sizes. It is interesting to highlight, how, for the butterfly dataset, the two relevant features $X_1$, $X_2$ are identified as the ones having the smallest bandwidths independently from the number of training points. 

\begin{figure}[h] 
\centering\includegraphics[width=1\linewidth]{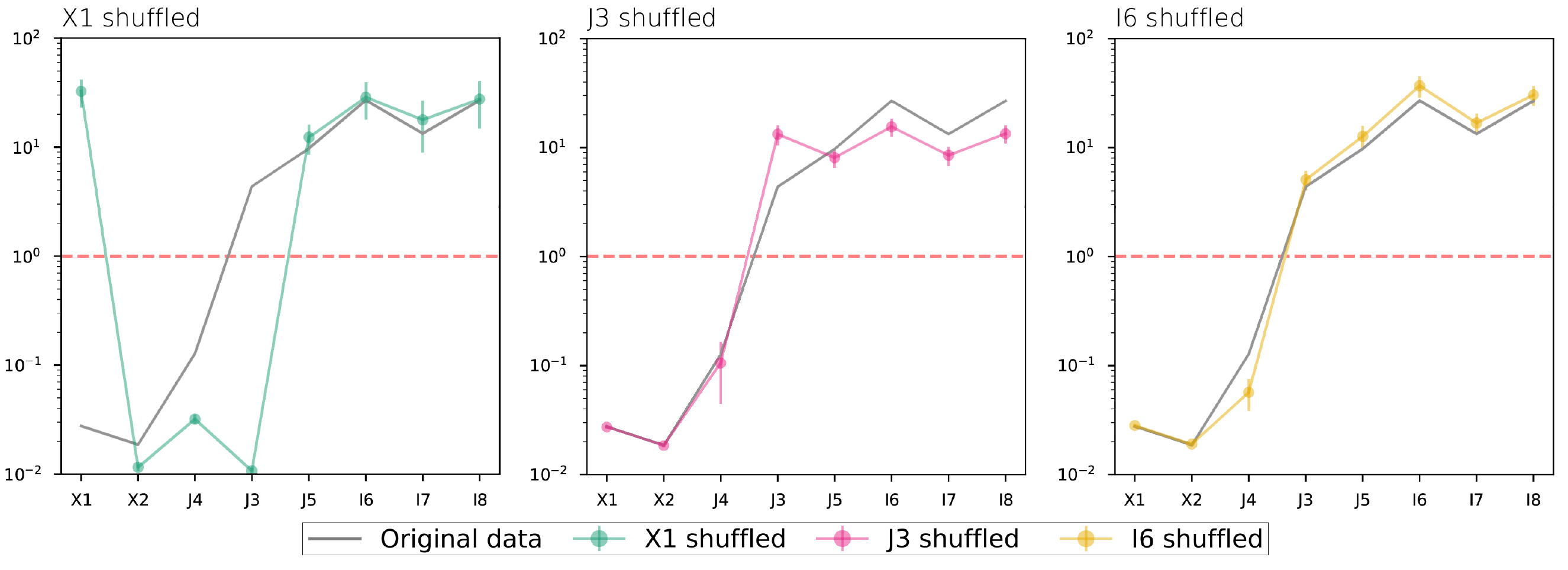}
\caption{Mean and  95\% confidence interval for the optimal bandwidth values after shuffling a relevant, a redundant and an irrelevant feature on a butterfly dataset having $n=2000$.}
\label{fig:sigma_shuff}
\end{figure}

Another way to study the importance of the features, like in random forest models, is to compare the results between original and shuffled features. Therefore, we analyzed the behaviour of the $\sigma_{opt}$ when the structure of one of the variables is destroyed by shuffling. Figure \ref{fig:sigma_shuff} shows the results of the 100 applications of AS after the shuffling of a relevant, a redundant and an irrelevant feature. Clearly,  when a relevant feature is shuffled, the corresponding value of the bandwidth surpasses the fixed threshold while the values of the redundant features, specifically $J_3$ and $J_4$, tend to decrease. Differently, when the irrelevant feature $I_6$ is shuffled, almost no changes are reported in the optimal bandwidths.

\subsection{Benchmark data}

The performances of AS have been compared with other well-known FS algorithms. Specifically, we tested all the selected algorithms on four datasets. The first one is a Friedman dataset \cite{friedman1991multivariate} generated with 30 features of which only 5 are relevant.
The other datasets have been downloaded from the UCI Machine Learning Repository (http://archive.ics.uci.edu/ml/index.php). Specifically we used the following datasets: Breast Cancer ($n=198$, $d=31$), California Housing ($n=2064$, $d=8$), Diabetes ($n=442$, $d=10$).
Four comparison algorithms were tested: Ftest, Mutual Information (MI), Correlation-based Feature Selection (CFS), ReliefF. To compare the performances of the different algorithms we used an external evaluator model. The experiment was designed as follows. Firstly, a FS algorithm has been applied to select the subset of features to test. In the following, random forest (RF) has been used as a benchmark regressor. To execute RF, data were split into training (80\% of the observations) and testing (20\%). RF was trained performing a 5-fold cross validation. Finally, the trained model has been used to perform a prediction on the testing data points to compute the MSE. The RF modelling has been repeated 20 times to avoid abnormal results due to data splitting. The entire procedure has been repeated for each of the FS algorithm tested and for all the compared datasets. 

Table \ref{tab:comparison} presents the results of the comparison. It can be easily pointed out how the sets of features selected by AS provide mean errors in line or better than those resulting from the application of the other algorithms, including also ReliefF, which is usually considered as a standard in the benchmark of FS algorithms.

\section{Software availability}\label{software}

The algorithm for AS proposed in this paper has been implemented in Python. The latest version is available on PyPi as pyGRNN.

\begin{table}
\centering
\tiny
\setlength{\tabcolsep}{1.8pt}
\begin{tabular}{l|cc|cc|cc|cc|cc}
\toprule
\multirow{2}{*}{\textbf{Dataset}} &   \multicolumn{2}{|c|}{\textbf{Ftest}}  & \multicolumn{2}{|c|}{\textbf{MI}} & \multicolumn{2}{|c|}{\textbf{CFS}} &  \multicolumn{2}{|c|}{\textbf{ReliefF}} &  \multicolumn{2}{|c}{\textbf{AS}}\\
\cline{2-11}
& \textbf{\#} &  \textbf{MSE} & \textbf{\#} &  \textbf{MSE} & \textbf{\#} &  \textbf{MSE} & \textbf{\#} &  \textbf{MSE} & \textbf{\#} &  \textbf{MSE} \\
\midrule
Friedman & 10 & 0.02($\pm$0.001) & 10 & 0.03($\pm$0.001) & 6 & 0.03($\pm 8.01e-5$) & 10 & 0.002($\pm 8.36e-5$) & 6 & 0.004($\pm 5.51e-4$)\\
Breast	& 11 &	0.081($\pm$0.010) &	11 & 0.077($\pm$0.009) & 10 & 0.077($\pm$0.011) & 11 & 0.077($\pm$0.009) & 11 & 0.076($\pm$0.013)\\
Calhouse & 6 & 0.016($\pm$0.001) &	6 & 0.015($\pm$0.001) & 6 & 0.013($\pm$0.001) & 6 & 0.011($\pm$0.001)& 7 &	0.012($\pm$0.001)\\
Diabetes & 5 & 0.044($\pm$0.004) & 5 & 0.042($\pm$0.004) & 	8 &	0.032($\pm$0.003) &	5 &	0.032($\pm$0.003) & 8 &	0.033($\pm$0.003)\\
\bottomrule
\end{tabular}
\caption{Number of selected features, mean and standard deviation of the MSE over 20 RF run for all the tested algorithms and datasets. For Ftest and MI the number of selected features has been fixed equal to the number of features selected through the ReliefF.}
\label{tab:comparison}
\end{table}

\section{Conclusions}\label{conclusions}

The AGRNN can be used to rank the features based on a distance/similarity criterion, and the definition of a proper threshold allows the recognition of relevant features. 
The main characteristics of the proposed FS approach have been studied using simulated datasets,  shuffling the features of the input space to destroy their relevancy to the output, if any. The proposed approach has then been compared with four other FS algorithms on four real world datasets downloaded from the public accessible repositories. The performances appeared as good or better as other algorithms in all case studies.

Future studies will investigate the behaviour of AS in higher dimensional spaces, together with the possibility of using it to recognize features redundancy or to perform multitask learning.


\begin{footnotesize}

\bibliographystyle{unsrt}
\bibliography{references.bib}

\end{footnotesize}


\end{document}